# SAGE-FM: A lightweight and interpretable spatial transcriptomics foundation model

Xianghao Zhan[1,2,*], Jingyu Xu[2,3,*], Yuanning Zheng[1,2,3], Zinaida Good[2-5,#], Olivier Gevaert[1,2,5,#]

[1]Department of Biomedical Data Science, Stanford University, Stanford, CA 94305, USA.

[2]Division of Computational Medicine, Department of Medicine, Stanford University, Stanford, CA 94305, USA.

[3]Division of Immunology and Rheumatology, Department of Medicine, Stanford University, Stanford, CA 94305, USA.

[4]Parker Institute for Cancer Immunotherapy, Stanford University, Stanford, CA 94305, USA.

[5]Team PROMISE, Weill Cancer Hub West, Stanford University, Stanford, CA 94305, USA.

*Co-first author.

#Co-senior author.

Correspondence: Olivier Gevaert  ogevaert@stanford.edu

**Abstract:**

Spatial transcriptomics enables spatial gene expression profiling, motivating computational models that capture spatially conditioned regulatory relationships. We introduce SAGE-FM, a lightweight spatial transcriptomics foundation model based on graph convolutional networks (GCN) trained with a masked-central-spot prediction objective. Trained on 416 human Visium samples spanning 15 organs, SAGE-FM learns spatially coherent embeddings that recover masked genes robustly, with 91% of masked genes showing significant correlations ($p < 0.05$). The SAGE-FM generated embeddings outperform MOFA and spatial transcriptomics in unsupervised clustering and preservation of biological heterogeneity. SAGE-FM generalizes to downstream tasks, enabling 81% accuracy in pathologist-defined spot annotation in oropharyngeal squamous cell carcinoma and improving glioblastoma subtype prediction relative to MOFA. *In silico* perturbation experiments further show that the model captures directional ligand–receptor and upstream–downstream regulatory effects consistent with ground truth. These results demonstrate that simple, parameter-efficient GCNs can serve as biologically interpretable and spatially aware foundation models for large-scale spatial transcriptomics.

**Introduction**

Spatial transcriptomics (ST) technologies enable transcriptome-wide gene expression profiling while preserving the spatial architecture of tissues, offering a transformative view of cellular organization in health and disease [Ståhl et al., 2016]. Unlike bulk or single-cell sequencing, ST captures not only what genes are expressed but also where they are expressed within intact tissue sections. As a result, ST has been widely adopted to advance both fundamental and translational biology. Example applications include: (i) cell type annotation within native contexts [Asp et al., 2022; Marx, 2021]; (ii) mapping proximity-based interactions [Wu et al., 2022; Ji et al., 2021]; and (iii) discovering spatial biomarkers, therapeutic targets, and disease mechanisms [Kuppe et al., 2022; Asp et al., 2022]. Despite these advances, ST datasets are high-dimensional, sparse, and noisy, and integrating spatial coordinates with transcriptomic signals remains technically challenging [Zhu et al., 2023; Liu et al., 2023]. These complexities underscore the need for advanced computational methods capable of extracting robust, biologically meaningful representations from ST data.

As single-cell and spatial transcriptomic datasets grow exponentially, there is a growing demand for foundation models that learn generalizable biological representations analogous to large language models in NLP [Devlin et al., 2019; Brown et al., 2020] or vision transformers in computer vision [Gevaert et al. 2025; Zhang et al. 2023; Azad et al. 2023]. Such models promise transferable, interpretable, and robust embeddings for diverse downstream biological tasks. Recent advances illustrate this potential. scGPT trains transformer models on millions of single-cell transcriptomes using masked gene modeling and contextual learning to capture broad gene–gene dependencies [Cui et al., 2024]. UCE (Universal Cell Embedding) integrates gene expression and protein-based embeddings to produce representations that generalize across tasks and modalities [Rosen et al., 2023]. Geneformer leverages protein–protein interaction networks and gene co-expression graphs to pre-train transformer models on large-scale transcriptomics data [Theodoris et al., 2023]. Most existing biological foundation models are non-spatial and trained on dissociated single-cell data, overlooking tissue organization, which plays a critical role in disease progression, tumor microenvironment dynamics, and therapeutic response. This motivates the development of spatially aware foundation models to capture the interplay between molecular and spatial contexts.

To develop foundation models and acquire biologically meaningful cell/spot embeddings, various computational strategies have been proposed to learn compact and informative representations from high-dimensional transcriptomic data, each with distinct theoretical underpinnings and trade-offs. First, reconstruction-based models [Lopez et al., 2018; Zhong et al., 2024; Dong & Zhang et al., 2022] aim to preserve variance across samples, batches, and modalities, enabling generative capabilities, modality transfer, and imputation. Second, matrix decomposition methods [Argelaguet et al., 2018; Argelaguet et al., 2020; Singh et al., 2019; Zhong et al., 2024] also preserve variance but provide interpretability and lower-dimensional factors, albeit with weaker generative and imputation abilities. Additionally, masked language modeling (MLM) approaches (e.g. scGPT; UCE; SPACE-GM) [Cui et al., 2024; Rosen et al., 2023; Fu et al., 2023; Wu et al., 2022] learn gene–gene regulatory relationships by predicting masked gene expression from context, offering high scalability and model expressiveness to produce powerful embeddings. Furthermore, contrastive learning methods (e.g., UCE, DeepST, NEST) [Rosen et al., 2023; Long et al., 2022; Zohora et al., 2024] learn neighborhood relationships and spatial patterns, excelling at spatial structure preservation but often requiring paired samples and extensive augmentation.

Among these, masked gene modeling offers key advantages for spatial foundation models [Baek et al., 2025]. By focusing on predicting missing genes from local neighborhood context, MLM encourages models to learn regulatory network structure rather than simple variance patterns, and it scales effectively to large datasets for robust embedding learning. This makes MLM especially suitable for pre-training spatial transcriptomics foundation models, where accurate modeling of local gene–gene dependencies is essential for generalization across tissues and experimental conditions.

Recent advances, such as the Cellular Interaction Foundation Model (CI-FM) [You et al., 2025], have demonstrated the promise of large-scale pretraining for spatial transcriptomics, showing that self-supervised approaches can effectively learn context-dependent gene expression patterns from millions of cells across diverse platforms. At the same time, most existing masked modeling strategies have been developed primarily for dissociated single-cell datasets and have not fully explored how localized gene masking can be applied in spatial transcriptomics. Building on these developments, we introduce a complementary approach that leverages masked gene prediction with spatially organized neighborhoods to jointly capture gene regulatory relationships and spatial context, thereby aiming to enable generalizable and biologically meaningful representation learning across diverse tissues.

While masked-gene-based pretraining has shown strong potential, most existing efforts fall into two categories. The first focuses on dissociated single-cell datasets [Brown et al., 2020], where models are evaluated mainly through gene-level reconstruction accuracy and lack assessment of spatial or tissue-level biological relevance [Madhu et al., 2025]. The second includes recent spatial transcriptomics models, which begin to incorporate spatial context but often remain limited to quantitative metrics without sufficient biological visualization or spatial validation of learned neighborhood patterns. These gaps underscore the need for a spatial transcriptomics foundation model that not only leverages masked gene prediction within local neighborhoods but is also evaluated on biologically interpretable, spatially grounded tasks to demonstrate mechanistic and tissue-level understanding.

Here, we introduce SAGE-FM (Spatial Analysis with Graph Embedding Foundation Model), a prototypical spatial transcriptomics foundation model based on a simple graph convolutional network (GCN) trained with a masked-gene prediction objective on 10x Visium neighborhoods (15-spot subgraphs). The model learns spatially conditioned gene-gene dependencies while remaining lightweight, interpretable, and parameter-efficient, avoiding the complexity of graph transformers or hierarchical architectures. Unlike prior works focused mainly on reconstruction quality, our evaluation emphasizes biological interpretability and spatial relevance through: (i) unsupervised spot clustering that preserves tissue and sample geometry; (ii) clinically meaningful spot annotation in oral squamous cell carcinoma (OSCC) and glioblastoma (GBM) with spatially coherent predictions, and (iii) *in silico* ligand–receptor perturbations that simulate directional gene regulatory effects. We pre-trained and validated our model on spatial transcriptomic datasets generated using the 10x Genomics Visium platform, a widely adopted whole-transcriptome technology that integrates gene expression profiling with histological context, providing a consistent and reproducible basis for cross-tissue benchmarking. Methodologically, our findings demonstrate that model simplicity does not preclude biological utility, a well-structured GCN with principled masking and spatial context achieves robust, interpretable, and computationally efficient performance, making it practical for broad adoption in biologically focused research.

As we develop and evaluate SAGE-FM, we include MOFA (Multi-Omics Factor Analysis) as our primary non–deep learning baseline because it tackles the same problem setting, which is the unsupervised representation learning from expression matrices, without using labels or spatial coordinates [Argelaguet et al., 2018; Argelaguet et al., 2020]. Its probabilistic factors are interpretable and widely adopted for summarizing variation across tissues and cohorts, providing a credible benchmark for downstream tasks such as region/cell-type annotation, gene imputation, and patient stratification; MOFA has also proven effective in biomedical discovery across genomics, transcriptomics, and proteomics applications [Gonçalves et al., 2022; Pekayvaz et al., 2024]. MOFA is also robust to missing values and scaling, trains with few hyperparameters, and is platform-agnostic, which reduces tuning advantages and makes comparisons reproducible. At the same time, its (approximately) linear factorization and lack of neighborhood encoding limit performance on spatially local phenomena, which is precisely what our masked, spatially aware objective is designed to capture.

**Results**

**SAGE-FM Effectively Predicts Masked Gene Expression and Imputes Missing Genes**

For pre-training, we curated 416 distinct human Visium spatial transcriptomics samples from the HEST1k dataset [Jaume et al., 2024], applying filters to ensure consistency (human-derived samples only, >10,000 measured genes per sample, and spot-level resolution). This yielded 981,797 spots across 14,558 genes, partitioned into training (332 samples, 776,171 spots), validation (42 samples, 103,353 spots), and test (42 samples, 102,273 spots) cohorts (**Fig. 1**).

From these samples, we constructed a 15-spot subgraph consisting of the spot itself (central node) and its 14 nearest spatial neighbors, defining a local neighborhood with a radius of approximately 500 μm. Each subgraph was modeled using a graph convolutional network (GCN), where nodes represented the 14,558-dimensional gene expression vector of each spot and edges encoded their physical distances. During pre-training, 20% or 30% of the central spot's genes were randomly masked, and the GCN was trained to predict these values using the remaining genes and neighborhood context (**Fig. S1**). Model performance was optimized using a masked MSE loss function and assessed using RMSE, mean $R^2$, median $R^2$ across samples, and the gene-wise Pearson correlation coefficient.

The best-performing model (30% masking, five-layer GCN with hidden units 1024–512–512–512–1024, trained for 126 epochs) achieved strong performance. On the validation set, the model reached RMSE = 0.305 (baseline without imputation = 0.465), mean $R^2$ = 0.256, and median $R^2$ = 0.290. Results were consistent on the test set (RMSE = 0.305, baseline = 0.465; mean $R^2$ = 0.256; median $R^2$ = 0.291). These $R^2$ values correspond to Pearson correlations of 0.5-0.6 between predicted and ground-truth values. In addition, the pre-trained graph neural network (GNN) achieved gene-wise Pearson correlations with a mean r = 0.457 and median r = 0.490, with 90.94% (13,151 of 14,461) of masked genes showing significant correlations ($p < 0.05$) in the validation set, confirming that the GNN successfully leveraged spatial context for masked gene recovery (**Table S1**).

We further evaluated the robustness of SAGE-FM (30% masking) in a systematic missing gene imputation task, in which 10–90% of genes were masked across all spots to simulate platform-specific gene dropout (**Figs. S2a–c**). Performance degraded gradually with increasing missingness, with a critical threshold at ~60% missing genes, where the Pearson correlation remained above 0.4 ($R^2 > 0.16$), representing 80% of the performance on the pre-training task. Beyond this point, performance dropped below 80% of the pre-training baseline. Gene-wise Pearson r across gene missingness on the HEST1k test set showed that both mean and median correlations declined gradually with increasing gene missingness (**Fig. S2d**). These missing gene imputation results demonstrate that the GNN embeddings effectively capture spatially informed dependencies, maintaining predictive power even under significant information loss.

**SAGE-FM Enables Biologically Meaningful Spot Clustering and Biological Heterogeneity Preservation**

To test whether the hidden-layer embeddings derived from the pre-trained GNN models captured biologically meaningful variance, we performed unsupervised clustering of spots and compared the results against known sample and tissue labels. For this analysis, we used three types of input representations: (1) GNN embeddings from the masked gene prediction model, (2) MOFA embeddings as a widely used dimensionality reduction baseline with sparsity constraint[Argelaguet et al., 2018; Argelaguet et al., 2020], and (3) raw spatial transcriptomics profiles under three transformations (raw counts, log1p, and log1p with Z-score normalization). Spot clustering was performed using k-means, either directly or after PCA-based dimensionality reduction to 200 dimensions (except for MOFA, which already has low dimensionality).

To assess generalization across samples and studies, we compared clustering results against sample ID and tissue type annotations and found that GNN embeddings outperformed both MOFA and raw transcriptomics data. GNN embeddings achieved higher Adjusted Rand Index (ARI) and lower Davies–Bouldin Index (DBI) across both label sets (**Fig. 2a**). While MOFA achieved slightly higher Silhouette scores in tissue-level clustering, the overall metrics indicate that GNN embeddings better reflect the biological structures present in the data.

Beyond spot-level clustering against labels, we next asked whether each embedding preserves global inter-sample and tissue geometry: that is, the pattern of similarities among sample and tissue centroids. For each representation, we computed sample-wise and tissue-wise centroids and measured pairwise similarities using cosine distance. Using the spatial transcriptomics data as the reference, GNN embeddings yielded distributions that were consistently closer to the reference than MOFA embeddings (**Figs. 2b, S3**). In addition, neighborhood ranking analysis showed that GNN embeddings maintained same-tissue samples as closer neighbors than MOFA and were able to capture irregularities in the original data, such as the unusually large distances between eye and uterus samples, which MOFA embeddings failed to preserve (**Table S3**).

Finally, we examined whether the learned embeddings grouped samples in biologically interpretable ways. In a hierarchically clustered cosine distance heatmap (**Fig. 2c**), GNN embeddings produced tight clusters that were homogeneous by tissue of origin, with the closest groups (cosine distance < 0.023) exclusively containing samples from the same tissue. Similarly, t-SNE visualizations (**Fig. 2d**) showed that spots derived from the same tissue were consistently positioned closer together when represented by GNN embeddings compared to MOFA or raw transcriptomics features. Consistent with these findings, unsupervised clustering at the spot level further revealed tissue-specific transcriptional programs (**Fig. S5**): each cluster displayed distinct dominant tissue compositions and differentially expressed marker genes. These results confirm that GNN embeddings capture coherent, biologically interpretable tissue structures across diverse samples.

To sum up, these analyses demonstrate that GNN embeddings capture both spot-level and sample-level biological heterogeneity, outperforming MOFA in clustering accuracy and in preserving biologically meaningful inter-sample relationships.

**SAGE-FM Enables Accurate Spot Pathology Annotation in Head and Neck OSCC Spatial Transcriptomics**

We further evaluated the generalizability of the pre-trained GNN embeddings in a downstream application of stratifying pathologist annotations using a head and neck oral squamous cell carcinoma (OSCC) Visium dataset with detailed expert annotations [Arora et al., 2023]. The pathologist labels represent independent biological ground truth that is not derived from transcriptomics, making this task a stringent test of representational quality.

To simplify the highly imbalanced dataset of 11 annotation categories, we grouped labels into three biologically coherent classes: Tumor (SCC, Keratin), Stroma (Lymphocyte-Negative Stroma, Lymphocyte-Positive Stroma, Glandular Stroma), and Muscle (Muscle, Lymphocyte-Positive Muscle). The dataset was partitioned into 8 training/validation samples and 4 independent test samples, with class distributions shown in Methods. Models were trained using GNN embeddings, MOFA embeddings, or raw spatial transcriptomics (ST) features as input, and evaluated with both a random forest classifier and a three-layer neural network.

The GNN embeddings consistently enabled superior pathologist annotation classification performance (**Fig. 3a**). Across five standard metrics: accuracy, macro-averaged F1-score, micro-averaged F1-score, macro-averaged recall, and macro-averaged precision, the models trained on GNN embeddings significantly outperformed those trained on MOFA embeddings or raw ST profiles ($p < 0.001$). The violin plots demonstrate that GNN-derived features yield both higher median performance and lower variability, highlighting the robustness of the learned representations (**Table S4**).

We further examined classification outputs on two representative OSCC samples (**Figs. 3b and 3c**). Visual comparison of the predicted spot annotations against the ground-truth pathologist labels revealed that the GNN-based models produced spatial maps with markedly higher concordance to expert annotations than those generated using MOFA or raw ST features. In particular, the GNN embeddings more accurately delineated the complex boundaries between tumor, stromal, and muscle regions, capturing fine-grained transitions that were frequently blurred or misclassified by the baseline approaches. The outlined regions in the figures illustrate this improvement, emphasizing the biological and clinical relevance of GNN-derived embeddings for histopathological interpretation.

To sum up, these results demonstrate that pre-trained GNN embeddings provide a powerful and biologically meaningful feature space for spot-level annotation in OSCC, outperforming established baselines in both quantitative and qualitative evaluations.

**SAGE-FM Enables Dominant Cell Type Identification for Glioblastoma Spatial Transcriptomics**

We next tested the utility of the pre-trained GNN embeddings in identifying transcriptional subtypes in glioblastoma [Ravi et al., 2022; Ren et al., 2023; Zheng et al., 2023]. The ground-truth labels for each gene expression spot were derived from deconvolution-based estimates of spot-level subtype composition, which defined five major glioblastoma subtypes: neural progenitor-like (NPC-like), oligodendrocyte progenitor-like (OPC-like), astrocyte-like (AC-like), mesenchymal-like 1 (MES1), and mesenchymal-like 2 (MES2).

We first formulated the problem as a five-class classification task, training random forest classifiers using either GNN embeddings, MOFA embeddings, or raw ST features as inputs. GNN embeddings enabled significantly better performance than MOFA, with an accuracy of $0.603 \pm 0.004$ and a macro F1-score of $0.496 \pm 0.004$, compared to MOFA's markedly lower accuracy ($0.228 \pm 0.001$) and macro F1-score ($0.105 \pm 0.001$) (**Fig. 4a**). Although the GNN models did not surpass raw ST features (accuracy = $0.691 \pm 0.003$, macro F1 = $0.547 \pm 0.003$), which is as expected, given that the subtype labels were derived directly from transcriptomics data, they approached the reference baseline closely, indicating strong representational power and biological relevance.

To better capture the continuous heterogeneity of glioblastoma subtypes, we reformulated the problem as a probability regression task and trained neural networks using binary cross-entropy loss to predict the full subtype probability distribution for each spot. Raw spatial transcriptomics features again outperformed embeddings (**Fig. 4b**; BCE = 0.287, median Pearson r = 0.984), serving as the upper bound since labels were derived from ST-based deconvolution. The GNN embeddings, however, substantially outperformed MOFA. With a two-layer neural network, GNN embeddings achieved BCE = 0.498 and a median Pearson r = 0.678, whereas MOFA embeddings performed poorly (BCE = 0.920, median Pearson r = –0.370). Visualizations of subtype predictions in a representative test sample (**Fig. 4c**) illustrate that GNN embeddings recovered all five subtypes with spatially coherent patterns, whereas MOFA missed two subtypes and failed to reproduce biologically plausible spatial structure.

To assess whether the GNN embeddings capture biologically relevant structure without supervision, we performed unsupervised clustering on the GNN embeddings. Five clusters were identified with the optimal Silhouette score, each exhibiting distinct transcriptional programs reflective of known glioblastoma subtypes and microenvironmental niches. Cluster 0 highly expressed *CHI3L1*, *GFAP*, and *AQP4*, characteristic of reactive astrocytes and inflammatory niches. Cluster 1 highly expressed *EGFR* and *NES*, indicative of neural progenitor-like tumor cells, whereas Cluster 2 upregulated *CA9* and *LOX*, consistent with a mesenchymal, hypoxia-associated phenotype. Cluster 3 expressed oligodendrocyte lineage genes such as *PLP1*, *MBP*, and *OLIG1*, representing an OPC-like niche, while Cluster 4 was marked by *PDGFRB*, *CSPG4*, and *PECAM1*, defining an endothelial/perivascular compartment. Together, these findings demonstrate that the GNN-derived embeddings recapitulate canonical glioblastoma cell states and their spatially organized microenvironments, underscoring the biological interpretability of the model (**Table S5**).

These findings confirm that GNN embeddings generalize across diverse disease contexts, including glioblastoma, enabling both accurate subtype prediction and biologically coherent unsupervised stratification.

*In silico* **Perturbation Shows SAGE-FM Learned Gene Regulatory Networks**

To determine whether the pre-trained GNN models learned spatially informed gene regulatory networks, we conducted two complementary *in silico* perturbation analyses. The first focused on 34 ligand–receptor gene pairs ("GeneX"–"GeneXR") included among the 14,558 genes modeled by SAGE-FM. The second extended this framework to a curated set of genes with experimentally validated upstream–downstream regulatory relationships [Nobusada, T. *et al.*, 2015].

The first *in silico* perturbation was on 34 ligand-receptor gene pairs. These pairs were identified by full-text search of known ligand and receptor terms within the Ensembl gene list, and only pairs where both genes were present in the modeled spatial transcriptomics dataset were retained for analysis. For each pair, we artificially upregulated ligand expression in all 14 neighboring spots by setting their values to the observed minimum or maximum across samples in the validation and test sets. The corresponding receptor genes' expression in the central spot was then predicted, with all receptor values masked during imputation to prevent leakage. We established a ground-truth reference by computing Pearson correlations between ligand expression in neighboring spots and receptor expression in central spots across both the validation and test sets. Only ligand-receptor pairs with significant and consistent correlation directionality across both validation and test sets were retained for evaluation. The top panel shows the positively and negatively correlated pairs based on reference (**Fig. 5a**), while the bottom panel shows that the GNN model reproduced the expected effects under *in silico* perturbation for the majority of these pairs. Among 23 pairs with significant positive correlations, the model correctly predicted receptor upregulation in 18 cases (e.g., IL4-IL4R, IL6-IL6R) (**Fig. 5b**). Similarly, for the two pairs with consistent negative correlations (APOB-APOBR, GCG-GCGR), the model correctly predicted receptor downregulation in both.

To be specific, we visualized representative ligand-receptor examples with established biological relationships, in which upregulation in neighboring spots is expected to upregulate, downregulate, or have no effect on the central spot's gene expression. Canonical ligand-receptor pairs such as IL4-IL4R (Th2 cytokine signaling) and CSF1-CSF1R (macrophage proliferation) are biologically expected to co-vary positively, while LIF-LIFR has been implicated in autocrine cancer signaling [Jorgensen & de la Puente, 2022]. In contrast, the F2 (prothrombin)-PROC/PROCR (protein C pathway) pair represents opposite branches of the hemostatic system-F2 driving coagulation and PROC/PROCR mediating anticoagulant feedback that limits further thrombin generation [Conway, 2012]. The predicted receptor responses largely mirror these biologically expected directions and the correlation analysis reference (**Fig. 5c**), indicating that the model captures mechanistically meaningful gene-gene interactions (**Table S6**).

To extend our validation beyond pairwise ligand-receptor interactions ("GeneX"-"GeneXR"), we did the second experiment to assess whether the model learned a broader gene regulatory network. We selected input genes (ligands) with known downstream target genes based on established regulatory pathways (**Table S7**) [Nobusada, T. *et al.*, 2015]. We evaluated the consistency of predicted downstream gene responses by comparing the effect size ratios (% upregulated spots - % downregulated spots) on the validation and test sets. We observed strong concordance on the test and validation sets (Pearson $r > 0.98$, $p < 0.001$), with input genes showing consistent directional effects on their downstream genes across both datasets.

To determine whether the observed downstream gene responses reflect specific regulatory relationships rather than general model behavior over all genes, we compared effect sizes between known downstream genes and randomly selected baseline control genes. For each input gene, we selected an equal number of randomly chosen genes (excluding the input gene and its known downstream targets) and performed identical perturbation experiments. Across 10 independent baseline replicates with different random seeds, downstream target genes consistently showed higher effect sizes than baseline control genes in both the test and validation sets. This significantly differential response pattern was observed in 7 out of 10 baseline replicates in both datasets ($p < 0.05$ for individual comparisons), demonstrating that the model's perturbation responses are specific to known regulatory relationships rather than reflecting general prediction biases. These results indicate that the pre-trained GNN embeddings capture biologically meaningful multi-gene regulatory network dependencies, extending beyond simple pairwise correlations to encode directional regulatory influences within spatial neighborhoods.

# Discussion

In this study, we developed and systematically validated SAGE-FM, a spatial transcriptomics foundation model based on a graph convolutional network (GCN) trained with a masked-gene prediction objective. This training strategy enables the GCN to learn spatially conditioned gene–gene dependencies from 10x Visium data, producing interpretable, spatially aware biological representations. Unlike most existing computational studies that emphasize reconstruction or contrastive objectives [Zhao et al., 2025; Madhu et al., 2025], our work places emphasis on biological interpretability and clinical relevance. Specifically, we validated the learned representations through a series of biologically meaningful tasks: (i) unsupervised spot clustering that preserves tissue and sample heterogeneity; (ii) two downstream tasks—spot annotation in glioblastoma and head and neck OSCC—which directly reflect clinical and pathological relevance; and (iii) *in silico* ligand–receptor perturbations that investigate spatial gene regulatory networks. These experiments go beyond gene-level evaluation and address the gap between computational performance metrics and biological interpretability. The findings demonstrate that SAGE-FM provides a biologically interpretable and computationally efficient foundation for spatial transcriptomics representation learning.

Our choice of a simple GCN over more complex graph neural networks (e.g. graph transformers) is worth elaborating upon. While recent biological foundation models often adopt hierarchical, attention-based, or transformer-style architectures [You et al., 2025; Zhao et al., 2025; Madhu et al., 2025], our results demonstrate that model simplicity and interpretability need not come at the expense of biological fidelity. A simple GCN effectively aggregates local spatial context, models continuous tissue geometry, and propagates molecular information across neighboring spots. Compared with transformer-based spatial models, GCNs are parameter-efficient, computationally lightweight, and easier to train and deploy. These are advantages for laboratories seeking practical, reproducible and lightweight computation tools. Moreover, GCNs offer transparent message-passing mechanisms that align naturally with the physical notion of inter-cellular signaling, which enable interpretability in ligand–receptor and neighborhood analyses.

We based our model development on the HEST1k dataset [Jaume et al., 2024], including only 10x Visium spatial transcriptomics data from human samples. Visium remains one of the most widely adopted and pathology-validated spatial transcriptomics platforms, characterized by its high sequencing depth, reliable gene coverage, and tissue-level resolution. This consistency in the training dataset enables robust masked-gene pretraining and rigorous benchmarking across tissues. Using a single, standardized platform avoids the confounding batch and resolution effects introduced by mixing technologies (e.g., Slide-seq, Stereo-seq, CosMx, MERFISH, Xenium) [Zhao et al., 2025; You et al., 2025; Madhu et al., 2025], ensuring that improvements in representation learning reflect biological generalization rather than platform heterogeneity.

Compared with prior studies, our framework introduces three key advances. First, while most existing spatial or multimodal foundation models focus on computational metrics such as reconstruction loss or clustering accuracy [Madhu et al., 2025], our study systematically evaluates biological interpretability through spatially and clinically grounded experiments. Many earlier works provide limited biological validation or omit spatial visualization altogether; in contrast, our results include spatial visualization, perturbation responses, and tissue-level coherence analyses that bridge molecular predictions to histopathological structures. Second, by training on well-curated Visium data and enforcing sample-level train/validation/test splits, we avoid data leakage, which is an oversight in some prior studies that partitioned regions within the same tissue slide into training, validation, and test sets. This ensures that model generalization is evaluated across truly unseen biological samples. Third, our framework establishes a benchmark protocol for biologically meaningful evaluation, including ligand–receptor *in silico* perturbation testing and cross-disease downstream validation, providing a reusable blueprint for future spatial foundation model development.

While our current framework demonstrates strong biological and clinical performance, several extensions could further enhance its scope and impact. First, applying the model to higher-resolution and multimodal spatial platforms, such as Stereo-seq, CosMx, or Xenium, would enable evaluation of its scalability to single-cell and subcellular spatial resolution. Integrating additional data modalities (e.g., proteomics, histopathology, or chromatin accessibility) could also help model more comprehensive spatial gene–protein interactions.

Second, the HEST1k–Visium benchmark established here provides a valuable testbed for future methodological development. More advanced architectures, such as graph transformers, heterophilic GNNs, and hierarchical message-passing networks, can be systematically compared using our biologically interpretable evaluation suite (including ligand–receptor perturbations and downstream clinical tasks) to assess their added value over lightweight GCNs.

Finally, expanding benchmarking across diverse tissues, species, and disease contexts will help evaluate model generalizability and identify universal versus tissue-specific spatial principles. Such cross-tissue generalization studies, coupled with integration into spatially resolved clinical pipelines, will move spatial foundation models toward translational and diagnostic applications in precision medicine.

More broadly, SAGE-FM provides a practical and extensible framework that can be readily adopted as a general-purpose feature extractor for new Visium spatial transcriptomics datasets without requiring labels. The learned embeddings can be directly used for downstream tasks such as spatial clustering, tissue annotation, patient stratification, and interrogation of spatial gene regulatory mechanisms. Owing to its simple and transparent GCN-based design, the framework also serves as a reusable foundation for future methodological extensions and integration into diverse experimental and clinical spatial transcriptomics pipelines.

## Methods

**Pre-training Dataset**

For the pre-training and initial evaluation of the spatial transcriptomics foundation model, we used the Human Expression Spatial Transcriptomics 1000 (HEST1k) dataset, a recently released large-scale machine learning resource for spatial transcriptomics research. HEST1k comprises over one thousand spatial transcriptomics samples collected across diverse tissues, species and disease contexts, to standardize benchmarking and enable foundation model development in this domain. For this study, to ensure consistency and sufficient coverage for model training, only human-derived Visium samples with more than 10,000 genes measured per spot were kept. After filtering, the dataset comprised 416 samples and 14,558 genes, which were divided into training (332, 80%), validation (42, 10%), and test (42, 10%) sets. The training/validation/test partition was based on samples (slides) instead of by spots/regions, to prevent data leakage.

**Model Architecture and Pre-training**

We used a graph convolutional network (GCN) architecture in which each spot was represented as a node (14,558-dimensional feature vector) and edges encoded Euclidean distances between spots. Subgraphs consisted of 15 spatially adjacent spots (~500 μm radius within Visium spatial transcriptomics), including a central spot, four closest neighbors on the same row, four closest neighbors on two adjacent rows, and one neighbor from each intermittent row.

During pre-training, 20% or 30% of genes in the central spot were randomly masked, and the model was trained to predict the masked values using both the unmasked genes of the central spot and the full gene expression profiles of the neighboring spots. The training objective was the masked mean squared error (MSE), calculated only on the masked genes of the central spots across all subgraphs.

We experimented with GCN architectures containing 4-5 layers and hidden unit sizes of 256, 512, 1024, and 2048, with ReLU activations in all hidden layers. The optimal configuration for each masking percentage was selected based on validation root mean squared error (RMSE) and is reported in the supplementary materials.

**Model Evaluation on Missing Gene Imputation**

To assess whether the foundation model learns the gene regulatory network, we first designed an evaluation task focused on systematic missing gene imputation. Between 10% and 90% of genes were randomly masked across all spots. The models trained with 20% and 30% masking strategies during pre-training were evaluated for their imputation ability. Performance was assessed using RMSE, mean and median $R^2$ across samples, mean and median gene-wise Pearson correlation coefficient $r$, and a critical threshold was defined as the percentage of missingness at which the mean $R^2$ across samples fell below 80% of the pre-training baseline performance. Performance was quantified by monitoring the change in predictive accuracy relative to the pre-training task. We observed a critical threshold at the point where the $R^2$ score dropped to 80% of its pre-training performance, defining when gene imputation became unreliable.

For this evaluation, we kept the best-performing models from the pre-training stage under the 20% and 30% masking strategies and applied them to the HEST1k test set. Among these, the model pre-trained with 30% masking demonstrated superior performance in imputing missing genes under increasingly severe dropout conditions. Consequently, this model was selected as the final foundation model for subsequent analyses, including the *in silico* perturbation study, model interpretation on the HEST1k validation and test sets, and two downstream biomedical machine learning tasks.

**Model Evaluation on Spot Clustering**

We evaluated clustering performance using GNN embeddings, MOFA embeddings, and raw spatial transcriptomics data under three transformations: raw counts, log1p, and log1p with Z-score normalization. Clustering was performed using K-means, either directly or following principal component analysis (PCA)-based dimensionality reduction to 200 components (MOFA embeddings were not coupled with PCA since they are already below 200 dimensions). Performance was evaluated against two ground-truth label sets: sample ID and tissue type. Metrics included the Adjusted Rand Index (ARI), Davies-Bouldin Index (DBI), and Silhouette score. This analysis provided an unsupervised perspective on whether the representations learned by the foundation model capture meaningful biological structure beyond what is available in raw expression profiles. Higher alignment of GNN embeddings with known labels suggests that the model has learned biologically structured representations, whereas poor alignment would indicate missing biologically relevant variance.

To further evaluate the biological interpretability of the GNN embeddings, we performed an additional unsupervised clustering analysis focused on identifying biologically coherent spot groups. Specifically, we varied the number of clusters and selected the clustering configuration that maximized the Silhouette score, thereby ensuring well-separated and internally cohesive clusters. Using this optimal clustering, we conducted one-versus-all differential expression (DEG) analyses to identify marker genes for each cluster and examined the tissue composition of the spots assigned to each cluster, with results shown in the Supplementary Materials. This analysis allowed us to determine whether clusters defined by GNN-derived embeddings corresponded to biologically distinct gene expression programs and tissue contexts, thereby providing a qualitative assessment of their biological meaningfulness.

**Model Evaluation on Biological Heterogeneity Preservation**

To assess whether the embeddings preserved global biological relationships, we computed sample-wise and tissue-wise centroids for each representation type. Pairwise cosine distance matrices were then calculated and compared with those derived from raw spatial transcriptomics profiles.

These cosine distance matrices were evaluated in two complementary ways. First, we normalized the distances to the range [0, 1] and treated the distance structure derived from the spatial transcriptomics data as the reference. We then computed the absolute error between each entry of the MOFA-based or GNN-based distance matrices and the corresponding entry in the spatial transcriptomics–based matrix. This allowed a direct quantitative comparison of how well each representation preserved the original biological relationships.

Second, we examined the relative neighborhood structure among samples belonging to the same tissue. For each sample, we ranked all other samples based on cosine distance from its centroid and recorded the average rank positions of same-tissue samples. Ideally, similar samples from the same tissue and same disease state should appear among each other's top nearest neighbors (average rank close to 1). This analysis was performed for all three representation types to assess how well they maintained the intra-tissue similarity and inter-tissue separation observed in the original data.

Together, these analyses evaluated whether the GNN embeddings captured the global tissue-tissue and sample-sample relationships present in the spatial transcriptomics data more accurately than conventional factor-based or raw-expression representations. Superior preservation of these biological distance relationships by GNN embeddings would indicate that the foundation model has learned a structured representation space aligned with true biological heterogeneity.

**Downstream Application Task 1: Head and Neck OSCC Pathologist Annotation**

We evaluated the generalizability and biological relevance of SAGE-FM in a downstream application of pathologist annotation prediction using a head and neck oral squamous cell carcinoma (OSCC) spatial transcriptomics dataset. This dataset, published by Arora et al. [Arora et al., 2023], comprises Visium spatial transcriptomics profiles of HPV-negative OSCC samples with expert pathologist annotations characterizing diverse tissue regions and microenvironments. The original study demonstrated that these spatial architectures, particularly the distinction between tumor core and leading edge, are strongly associated with patient survival and therapy response, underscoring the clinical significance of accurate spatial annotation in OSCC. Our goal was to determine whether the GNN embeddings generated by the pre-trained foundation model can serve as informative features for predicting pathologist-defined tissue categories, and to compare their predictive performance against MOFA embeddings and raw spatial transcriptomics profiles. This task offers a strong benchmark because the pathologist annotations represent independent biological labels not derived from the transcriptomics data itself, allowing an unbiased evaluation of the representational quality of spatial transcriptomics, MOFA embeddings, and GNN embeddings.

The dataset contains 11 original annotation categories (SCC: 15,395; Lymphocyte-Negative Stroma: 3,241; Lymphocyte-Positive Stroma: 2,806; Muscle: 1,190; Glandular Stroma: 566; Keratin: 526; Artifact: 520; Lymphocyte-Positive Muscle: 219; Non-cancerous Mucosa: 182; Cautery: 107; Fold: 25; Edge Effects: 16; Artery/Vein: 16; nan: 1,562). To simplify the task and address class imbalance, we consolidated these labels into three biologically coherent major classes: tumor (SCC, Keratin), stroma (Lymphocyte-Negative Stroma, Lymphocyte-Positive Stroma, Glandular Stroma), and muscle (Muscle, Lymphocyte-Positive Muscle).

The dataset was partitioned into 8 training/validation samples and 4 hold-out test samples. The training set contained 10,773 tumor, 4,266 stroma, and 1,160 muscle spots, while the test set contained 5,148 tumor, 2,347 stroma, and 249 muscle spots. To ensure comparable validation size with the test set during model selection, we performed two-fold cross-validation on the training/validation set.

Two types of classifiers were trained using each representation (GNN embeddings, MOFA embeddings, and raw spatial transcriptomics): a random forest classifier, and a three-layer fully connected neural network. The hyperparameters tuned for the RF classifier were the number of estimators and maximum depths while the number of units and training epochs of the DNN was tuned. These models were implemented with the scikit-learn package for simplicity. Model performance was evaluated on the hold-out test set using accuracy, macro-averaged F1-score, macro-averaged precision, and macro-averaged recall.

**Downstream Application Task 2: Glioblastoma Dominant Cell Type Identification**

To assess the applicability of the foundation model in transcriptional subtyping, we curated and assembled spatial transcriptomics datasets of glioblastoma from three public resources [Ravi et al., 2022; Ren et al., 2023]. Each spot had previously been annotated by deconvolution-based estimates of dominant cell type, which served as the ground truth labels [Zheng et al., 2023]. Five established glioblastoma subtypes were considered: astrocyte-like (AClike), oligodendrocyte progenitor cell-like (OPClike), neural progenitor cell-like (NPClike), mesenchymal subtype 1 (MESlike1), and mesenchymal subtype 2 (MESlike2). We applied the pre-trained GNN model to generate hidden-layer embeddings based on gene expression.

We first formulated this task as a five-class classification problem to directly predict the dominant subtype of each spot. Random forest classifiers were trained using either the GNN embeddings, MOFA embeddings, or raw spatial transcriptomics profiles as inputs. The dataset was partitioned into 15 training/validation samples and 4 hold-out test samples. Hyperparameters were tuned via four-fold cross-validation on the 15 training/validation samples. Model performance was evaluated on the hold-out test set using standard multi-class metrics, including accuracy, macro-averaged F1-score, macro-averaged precision, and macro-averaged recall, enabling a direct comparison of representation types.

To further evaluate whether the GNN embeddings better capture the continuous heterogeneity of subtype composition, we formulated an additional task to predict the deconvoluted subtype likelihoods of each spot. A feed-forward neural network was trained for the regression of the five subtype likelihoods using either GNN embeddings, MOFA embeddings, or raw spatial transcriptomics profiles as input features. The networks employed ReLU activations for hidden layers and a sigmoid activation at the output layer to produce probability estimates for the five subtypes. Cross-entropy loss was used as the objective. We tested two- and three-layer architectures, tuning the number of hidden units using four randomly held-out validation samples from the 15 training/validation set to accelerate model selection.

For evaluation, the predicted probabilities were first converted to discrete subtype predictions by taking the argmax across the five outputs, and the resulting classifications were assessed using the same performance metrics as the random forest models. Additionally, to assess the models' ability to recover the continuous likelihood structure, we calculated the mean and median Pearson correlation coefficients and the mean and median Spearman correlation coefficients between the predicted and ground-truth subtype likelihoods across spots. This comprehensive evaluation allowed us to compare how effectively GNN embeddings, MOFA embeddings, and raw expression data support subtype identification and likelihood estimation in glioblastoma.

Besides supervised clustering, we clustered Visium spots using the learned GNN embeddings without supervision. The embeddings were clustered using K-means with k ranging from 4 to 10. Cluster quality was assessed using the Silhouette score as the primary criterion alongside Calinski-Harabasz and Davies-Bouldin indices; k=5 maximized the Silhouette score and was selected for downstream analyses. To define cluster-specific markers, raw counts were normalized to 10,000 per spot and log-transformed to stabilize variance. We performed one-versus-rest differential testing for each cluster using the Wilcoxon rank-sum test across 23,872 expressed genes, declaring significance at FDR < 0.05 (Benjamini–Hochberg). This strategy identified both upregulated markers (enriched within a given cluster) and informative downregulated genes (depleted relative to the remainder), which were summarized in volcano plots and exported for pathway/functional annotation.

*In silico* **Perturbation**

To investigate whether the pre-trained GCN captures a biologically meaningful gene regulatory network, we performed two forms of *in silico* perturbation using local subgraphs from the HEST1k validation and test sets. The first form focused on ligand–receptor interactions. A total of 34 ligand–receptor gene pairs were selected for perturbation analysis. For each ligand, we computed the minimum and maximum log1p-transformed expression across the HEST1k validation and test sets, and perturbed neighboring-spot ligand expression by clamping values to these extrema, thereby simulating strong upregulation on data unseen during training. To prevent information leakage, expression values for all corresponding receptor genes in the central spots were masked prior to prediction. The model was then used to impute receptor expression in the central spots under the perturbed conditions. Predicted receptor responses were quantified as the proportion of central spots exhibiting upregulation, downregulation, or unchanged expression relative to baseline (threshold $10^{-8}$). As a ground-truth reference, we computed Pearson correlations between mean ligand expression in neighboring spots and receptor expression in central spots across the validation and test sets.

In parallel, we carried out a second form of perturbation aimed at evaluating whether the model learns broader upstream–downstream regulatory networks. For each ligand, we defined two perturbation conditions by clamping its log1p-transformed expression in all 14 neighboring spots to the 0th or 100th percentile of its observed values in the HEST1k validation and test sets, while masking all downstream target genes in the central spot. The model then imputed central-spot expression for each downstream gene under both perturbation conditions. For each ligand-target pair, we classified central spots as upregulated, downregulated, or unchanged using a detection threshold of $10^{-8}$ and summarized responses as an effect size ratio $E$, defined as:

$$E = \frac{N_{up} - N_{down}}{N_{total}}$$

Where $N_{up}$ denotes the number of central spots exhibiting upregulation of the target gene, $N_{down}$ denotes the number of central spots exhibiting downregulation, and $N_{total}$ represents the total number of spots evaluated. Mean effect size ratios were computed per input gene and used for subsequent concordance and control analyses. For baseline controls, we sampled an equal number of non-target genes (excluding the input gene and its known downstream targets), repeated this random sampling 10 times with different seeds, and applied the same perturbation and effect-size computation to obtain 10 baseline replicates.

**Statistical Tests**

For the *in silico* perturbation experiments, we evaluated the association between ligand expression in neighboring spots and receptor expression in central spots using the Pearson correlation coefficient. Correlations were computed separately on the HEST1k validation and test sets. Ligand–receptor pairs were considered biologically associated only if they exhibited statistically significant correlations (two-tailed) on both sets and showed consistent correlation directionality. Pairs that were non-significant on either set or showed discordant directionality between the validation and test sets were classified as undetermined relationships and excluded from further analysis.

For the two downstream application tasks, model performance comparisons were assessed using bootstrapping-based resampling and paired statistical tests to ensure result robustness and reproducibility. Specifically, we performed 30 bootstrap experiments on the training data to obtain distributions of the classification performance metrics (accuracy, macro-F1, macro-precision, macro-recall) for each model. Pairwise comparisons between models trained on GNN embeddings, MOFA embeddings, and raw spatial transcriptomics data were then conducted using Wilcoxon signed-rank tests, a non-parametric test appropriate for paired, non-normally distributed data. This framework enabled statistically robust comparisons of predictive performance across different feature representations. For the downstream perturbation experiments, two‑sample t‑tests were used to compare effect size ratios between downstream target genes and randomly selected baseline control genes, both pooled across all baseline replicates and for each replicate individually.

**Acknowledgements**

We thank the members of the Gevaert Laboratory and the Good Laboratory for their constructive feedback and helpful discussions. This work was in part supported by the National Institutes of Health (NIH) / National Cancer Institute (NCI) awards R01CA260271 (O.G.) and 2P01CA049605 (Z.G.), NIH/OD 1OT2OD038101 (O.G., Z.G.), an institutional pilot grant IRG-23-1074369-01-IRG from the American Cancer Society and the Stanford Cancer Institute (SCI), an NCI-designated Comprehensive Cancer Center (Z.G.), a Kona Innovation Challenge grant (C-04134) from the Parker Institute for Cancer Immunotherapy (PICI) (Z.G.), Weill Cancer Hub West: Team PROMISE (O.G., Z.G.). Z.G. was supported by the PICI Parker Bridge fellowship and the NIH/NCI Pathway to Independence award (1K99CA293149, 4R00CA293149). Y.Z. was supported by the NIH/NCI Pathway to Independence award (1K99CA293249). Z.G. is a member of the Parker Institute for Cancer Immunotherapy, which supports the Stanford University Cancer Immunotherapy Program. The content is solely the responsibility of the authors and does not necessarily represent the official views of the NIH.

**Author Contributions**

X.Z., J.X., Y.Z., O.G., and Z.G. conceived the study. X.Z., J.X., and Y.Z. developed the methodology. X.Z. and J.X. performed the investigations and visualizations. X.Z. and J.X. wrote the original draft. Y.Z., Z.G., and O.G. reviewed and edited the manuscript. Z.G. and O.G. supervised the project.

**Declaration of interests**

Z.G. is an inventor on three patent applications, holds equity and advises Boom Capital Ventures, and received reagents, technical support, and/or speaker fees from 10x Genomics, Standard Biotools, AstraZeneca, and Sangamo Therapeutics. None of the above interests were related to the research described in this manuscript. All other authors declare no competing interests.

**Code Availability**

The data preprocessing (subgraph generation), pre-training code, evaluation on the pre-training task, missing gene imputation code, *in silico* perturbation code, code to run post-training inference and obtain embeddings for new data are shown in: https://github.com/dayenai/SAGE-FM/tree/main

**Data Availability**

The HEST-1k dataset used for model pre-training was obtained from the HEST-1k resource described by Jaume et al.; HEST-1k, together with the HEST-Library and HEST-Benchmark, can be accessed via its GitHub repository: https://github.com/mahmoodlab/HEST.

For downstream task 1 (OSCC pathologist annotation), we used the HPV-negative oral squamous cell carcinoma Visium dataset from Arora et al. (Nat Commun 2023). The raw and SpaceRanger-processed spatial transcriptomics data are deposited in the NCBI Gene Expression Omnibus (GEO) under accession GSE208253, and processed Seurat objects and loom files are available on Figshare (https://doi.org/10.6084/m9.figshare.20,304456.v1) and through the public OSCC portal (http://www.pboselab.ca/spatial_OSCC/).

For downstream task 2 (glioblastoma subtyping), we curated spatial transcriptomics datasets from three published glioma studies. The multiregion glioblastoma multi-omics dataset from Ravi et al. is available in Dryad under DOI 10.5061/dryad.h70rxwdmj. The malignant glioma spatial atlas from Ren et al. is deposited with raw sequencing in the Genome Sequence Archive for Human (GSA-Human; HRA001865, HRA001960, HRA003511) and processed spatial transcriptomics data in GEO under accession GSE194329, with additional processed matrices on Figshare (https://doi.org/10.6084/m9.figshare.20653908). We also used the compiled glioblastoma spatial datasets from Zheng et al., who integrated publicly available Visium and related datasets; these can be accessed via the accession URLs listed in Supplementary Data 1 of their study.

**References**

1 Ståhl, P. L. *et al.* Visualization and analysis of gene expression in tissue sections by spatial transcriptomics. *Science* **353**, 78–82 (2016).

2 Asp, M. *et al.* A spatiotemporal organ-wide gene expression and cell atlas of the developing human heart. *Nat. Methods* **19**, 1353–1364 (2022).

3 Marx, V. Method of the Year: spatially resolved transcriptomics. *Nat. Biotechnol.* **39**, 117–121 (2021).

4 Wu, R. *et al.* Comprehensive single-cell spatial transcriptomic profiling of mouse organogenesis. *Nature* **601**, 131–136 (2022).

5 Ji, A. L. *et al.* Multimodal analysis of composition and spatial architecture in human squamous cell carcinoma. *Nature* **598**, 571–576 (2021).

6 Kuppe, C. *et al.* Spatial multi-omic map of human myocardial infarction. *Nature* **608**, 766–777 (2022).

7 Asp, M. *et al.* [ZG1] A spatiotemporal organ-wide gene expression and cell atlas of the developing human heart. *Nat. Methods* **19**, 1353–1364 (2022).

8 Zhu, Q. *et al.* Integration of spatial and single-cell transcriptomic data elucidates mouse organogenesis. *Nat. Biotechnol.* **41**, 592–603 (2023).

9 Liu, Y. *et al.* High-spatial-resolution multi-omics sequencing via deterministic barcoding in tissue. *Nat. Methods* **20**, 1364–1375 (2023).

10 Devlin, J., Chang, M.-W., Lee, K. & Toutanova, K. BERT: Pre-training of deep bidirectional transformers for language understanding. *Proc. NAACL-HLT* 1, 4171–4186 (2019).

11 Brown, T. B. *et al.* Language models are few-shot learners. *Adv. Neural Inf. Process. Syst.* 33, 1877–1901 (2020).10 Cui, H., Wang, C., Maan, H. *et al.* scGPT: toward building a foundation model for single-cell multi-omics using generative AI. *Nat Methods* 21, 1470–1480 (2024).

12 Rosen, Y. *et al.* Universal Cell Embeddings: A foundation model for cell biology. *bioRxiv* (2023).

13 Theodoris, C. V. *et al.* Transfer learning enables predictions in network biology. *Nature* **613**, 684–692 (2023).

14 Lopez, R. *et al.* Deep generative modeling for single-cell transcriptomics. *Nat. Methods* **15**, 1053–1058 (2018).

15 Zhong, C. *et al.* Interpretable spatially aware dimension reduction of spatial transcriptomics with STAMP. *Nat. Methods* **21**, 2072–2083 (2024).

16 Dong, K. & Zhang, S. *et al.* Deciphering spatial domains from spatially resolved transcriptomics with adaptive graph attention auto-encoder (STAGATE). *Nat. Commun.* **13**, 5791 (2022).

17 Argelaguet, R. *et al.* Multi-Omics Factor Analysis—a framework for unsupervised integration of multi-omics data sets. *Mol. Syst. Biol.* **14**, e8124 (2018).

18 Argelaguet, R. *et al.* MOFA+: a statistical framework for comprehensive integration of multi-modal single-cell data. *Genome Biol.* **21**, 111 (2020).

19 Gonçalves, E. *et al.* Pan-cancer proteomic map of 949 human cell lines. *Cell Syst.* **13**, 965–982.e8 (2022).

20 Pekayvaz, K. *et al.* Multiomic analyses uncover immunological signatures in acute and chronic coronary syndromes. *Nat. Med.* **30**, 1696–1710 (2024).

21 Singh, A. *et al.* DIABLO: an integrative approach for identifying key molecular drivers from multi-omics assays. *Bioinformatics* **35**, 3055–3062 (2019).

22 Zhong, C[ZG1] ., Ang, K. S. & Chen, J. Interpretable spatially aware dimension reduction of spatial transcriptomics with STAMP. *Nat. Methods* **21**, 2072–2083 (2024).

23 Rosen, Y. *et al.* [ZG2] Universal Cell Embeddings: A Foundation Model for Cell Biology. *bioRxiv* (2023).

24 Wu, Z. *et al.* Graph deep learning for the characterization of tumour microenvironments from spatial protein profiles in tissue specimens. *Nat. Biomed. Eng.* **6**, 1435–1448 (2022).

25 Wu, Z. *et al.* Identifying spatial cellular structures with SPACE-GM. *Nat. Rev. Cancer* **23**, 508 (2023).

26 Fu, X., Mo, S., Shao, A., Laurent, A., Buendia, A., Ferrando, A. A., Ciccia, A., Lan, Y., Palomero, T., Owens, D. M., Xing, E. P. & Rabadan, R. GET: a foundation model of transcription across human cell types. *bioRxiv* (2023).

27 Long, Y. *et al.* DeepST: A versatile graph contrastive learning framework for spatial transcriptomics. *bioRxiv* (2022). https://doi.org/10.1101/2022.08.02.502407

28 Zohora, F. T. *et al.* NEST: Spatially-mapped cell-cell communication patterns (NEural network on Spatial Transcriptomics). *bioRxiv* (2024). https://doi.org/10.1101/2024.03.19.585796

29 Baek, S., Song, K. & Lee, I. Single-cell foundation models: bringing artificial intelligence into cell biology. *Exp. Mol. Med.* **57**, 1–? (2025). https://doi.org/10.1038/s12276-025-01547-5

30 You, Y., Wang, Z., Fleisher, K., Liu, R. & Thomson, M. Building foundation models to characterize cellular interactions via geometric self-supervised learning on spatial genomics (CI-FM). *bioRxiv* (2025).

31 Nobusada, T. *et al.* Update of the FANTOM web resource: enhancement for studying noncoding genomes. *Nucleic Acids Res.* **53**, D419–D424 (2025).

32 Jorgensen, M. M. & de la Puente, P. Leukemia inhibitory factor: an important cytokine in pathologies and cancer. *Biomolecules* **12**, 217 (2022).

33 Conway, E. M. Mechanisms of anticoagulant and cytoprotective actions of the protein C pathway. *Front. Biosci.* **17**, 1006–1027 (2012).

34 Arora, R. *et al.* Spatial transcriptomics reveals distinct and conserved tumor core and edge architectures that predict survival and targeted therapy response. *Nat. Commun.* **14**, 5095 (2023).

35 Ravi, V. M. *et al.* Spatially resolved multi-omics deciphers bidirectional tumor–host interdependence in glioblastoma. *Cancer Cell* **40**, 639–655.e13 (2022).

36 Ren, Y. *et al.* Spatial transcriptomics reveals niche-specific enrichment and vulnerabilities of radial glial stem-like cells in malignant gliomas. *Nat. Commun.* **14**, 1028 (2023)

37 Zheng, Y., Carrillo-Perez, F., Pizurica, M., Heiland, D. H. & Gevaert, O. Spatial cellular architecture predicts prognosis in glioblastoma. *Nat. Commun.* **14**, 4122 (2023) 38 Zhao, S., Luo, Y., Yang, G., Zhong, Y., Zhou, H., & Nie, Z. (2025). SToFM: A Multi-scale Foundation Model for Spatial Transcriptomics. In Proceedings of the 42nd International Conference on Machine Learning (ICML 2025), PMLR 267, Vancouver, Canada.

39 Madhu, H., Rocha, J. F., Huang, T., Viswanath, S., Krishnaswamy, S., & Ying, R. (2025). HEIST: A Graph Foundation Model for Spatial Transcriptomics and Proteomics Data. bioRxiv preprint, arXiv:2506.11152v2.

40 You, Y., Wang, Z., Fleisher, K., Liu, R., & Thomson, M. (2025). *Building foundation models to characterize cellular interactions via geometric self-supervised learning on spatial genomics (CI-FM). bioRxiv* preprint, doi:[10.1101/2025.01.25.634867].

41 Jaume, G., Doucet, P., Song, A. H., Lu, M. Y., Almagro-Pérez, C., Wagner, S. J., Vaidya, A. J., Chen, R. J., Williamson, D. F. K., Kim, A., & Mahmood, F. (2024). *HEST-1k: A Dataset for Spatial Transcriptomics and Histology Image Analysis*. arXiv preprint arXiv:2406.16192.

# Figures

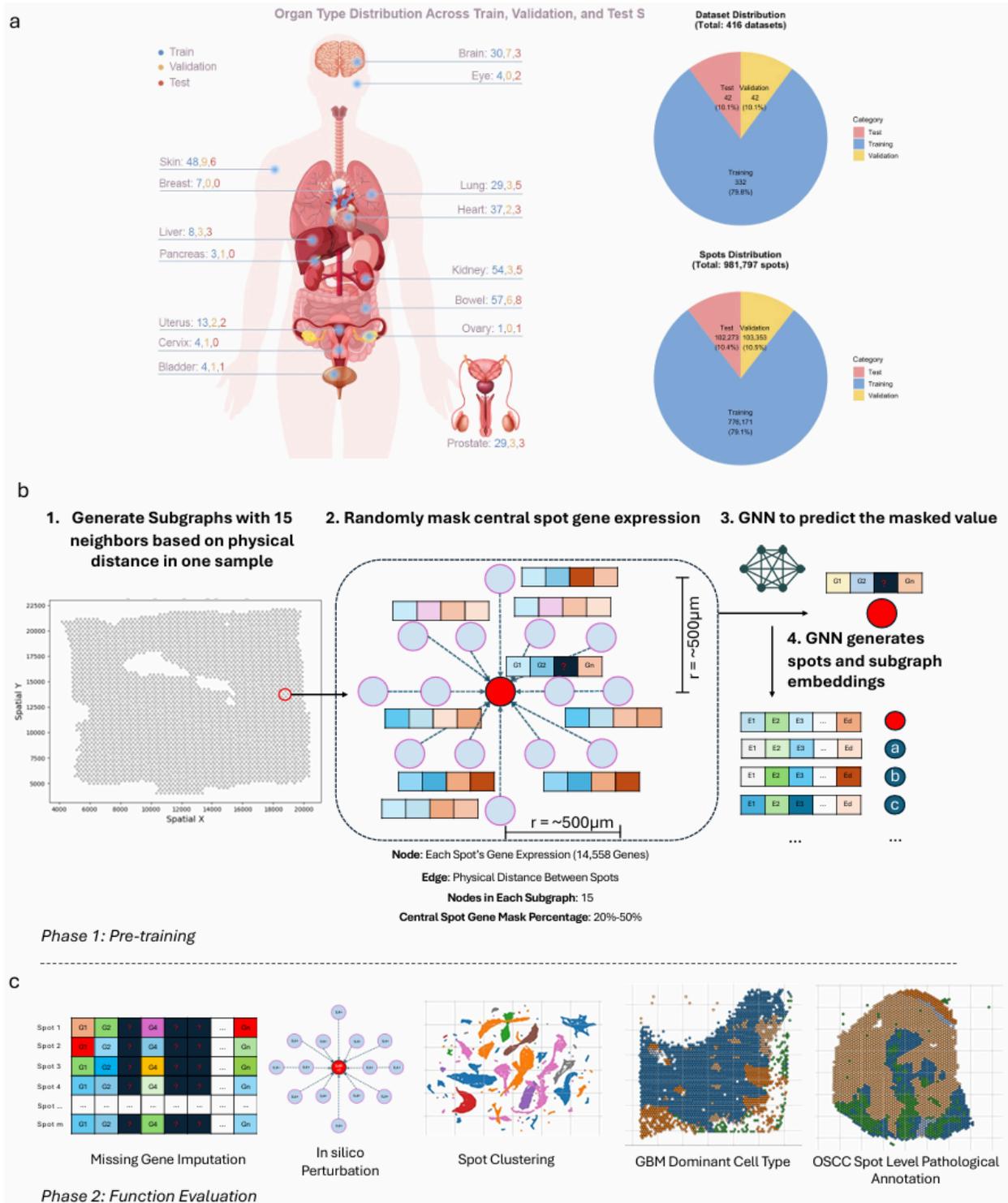

**Figure 1. The flowchart of the development and evaluation of SAGE-FM. (a)** A summary of the pre-training dataset of 416 human samples across 15 organs. The samples are partitioned randomly into 80% training, 10% validation, and 10% test samples. **(b)** An illustration of the pre-training method to train

a Graph Neural Network by creating 15-spot subgraphs, masking genes in the central spot, and predicting the missing gene expression in the central spot to learn spatial gene regulatory network and spatial transcriptomics embeddings. **(c)** The comprehensive model evaluation pipeline through a series of biologically interpretable downstream tasks. The evaluation ranges from fundamental functions like gene imputation, *in silico* perturbation, unsupervised spot clustering, to complex, clinically relevant applications like predicting expert pathologist annotations in OSCC and identifying dominant cell types and cancer subtypes in GBM.

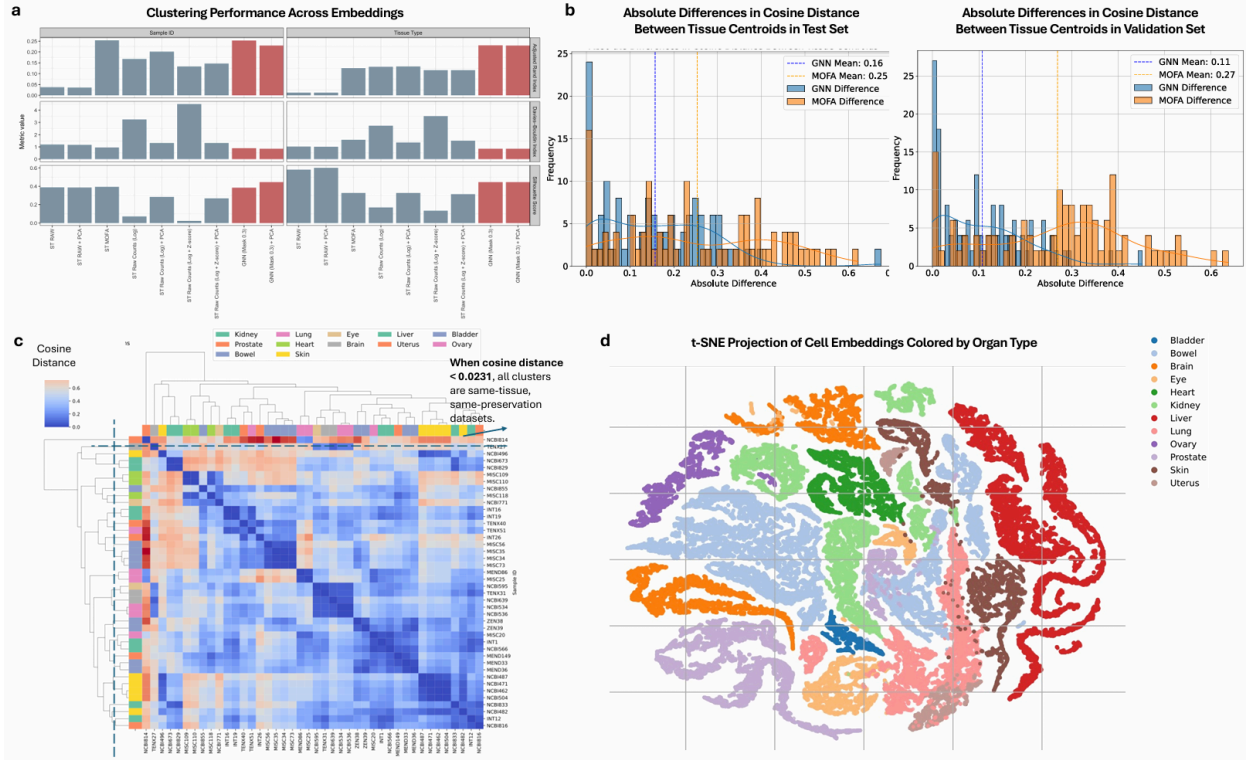

**Figure 2. Clustering performance and biological interpretability of GNN-derived embeddings.**

**(a)** Bar plots of unsupervised clustering performance with different embeddings. Unsupervised spot clustering was performed using GNN embeddings, MOFA embeddings, and raw spatial transcriptomics data under three transformations (raw counts, log1p, and log1p with Z-score transformation), with or without PCA dimensionality reduction. Clustering results evaluated against sample ID and tissue annotations show that GNN embeddings achieve higher Adjusted Rand Index (ARI) and lower Davies–Bouldin Index (DBI), with competitive Silhouette scores.

**(b)** Distribution density plots showing preservation of global inter-sample relationships. Cosine distance matrices from tissue-wise centroids from GNN embeddings and MOFA embeddings were computed and compared against the reference cosine distance matrix from spatial transcriptomics. The distribution plots show that GNN embeddings yield lower mean absolute to the reference than MOFA embeddings, indicating better preservation of biological heterogeneity.

**(c)** Cosine distance heatmap with hierarchical clustering based on 12 tissues' centroids using GNN embeddings. Hierarchical clustering reveals tight, homogeneous clusters by tissue of origin, with the lowest-level clusters (cosine distance < 0.0231) containing same-tissue exclusively.

**(d)** t-SNE visualization of spot embeddings, showing that GNN-derived embeddings group spots from the same tissue in close proximity, highlighting biological interpretability.

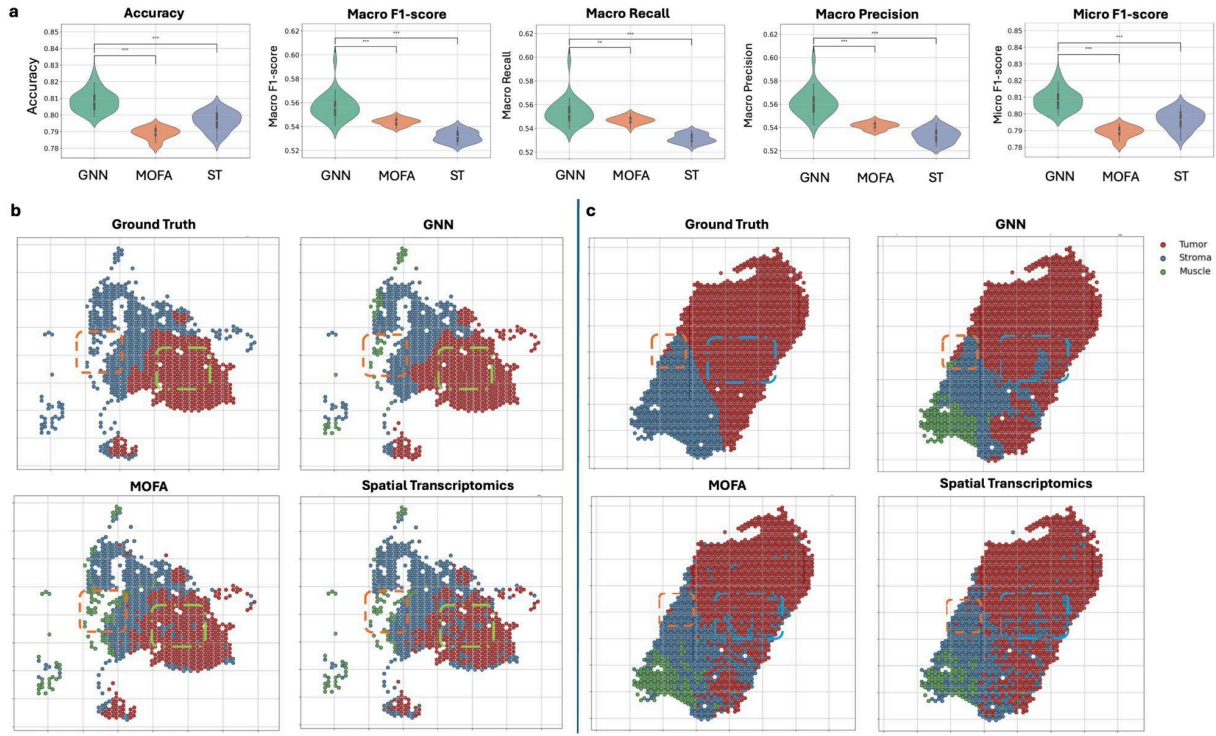

**Figure 3. Evaluation of GNN-derived embeddings for pathological annotation prediction in OSCC.**
a) Violin plots comparing classification performance metrics across best classification models based on different embeddings. Models trained with GNN embeddings (random forest), MOFA embeddings (random forest), or raw spatial transcriptomics (ST) (three-layer neural network) were evaluated on an oral squamous cell carcinoma (OSCC) dataset annotated by pathologists. Across five evaluation metrics (accuracy, macro F1-score, micro F1-score, macro recall, and macro precision), GNN embeddings significantly outperformed MOFA embeddings and ST features in all five metrics. b–c) Two representative examples showing predicted versus ground truth pathological annotations. Predictions generated with GNN embeddings displayed higher concordance with ground truth labels than MOFA or ST features. In particular, GNN-based models formed more solid and consistent boundaries between tumor, stromal, and muscle regions, rather than misassigning stromal spots within dense tumor regions as observed with baseline embeddings.

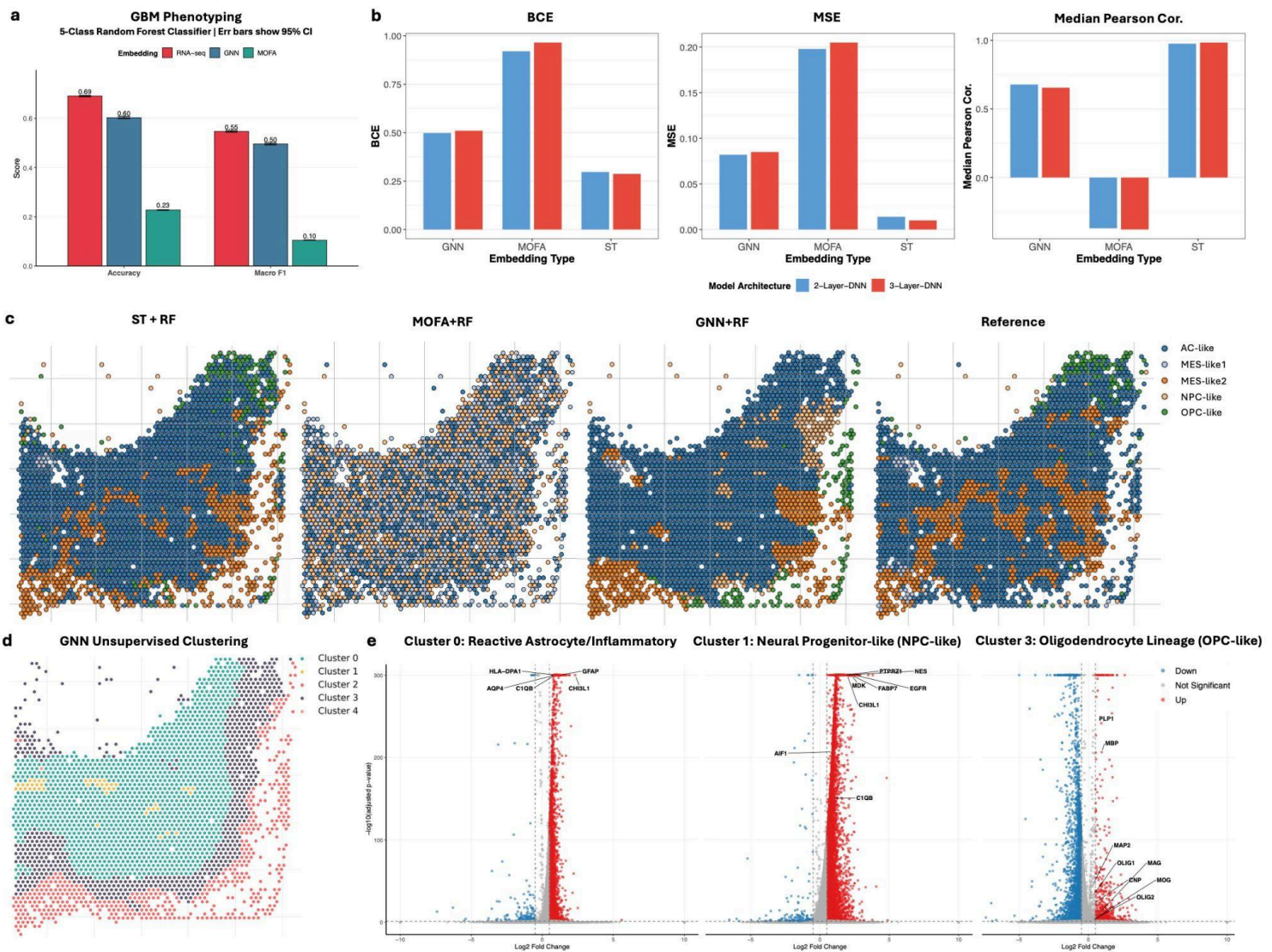

**Figure 4. Glioblastoma subtype identification using GNN-derived embeddings.** a) Five-class classification performance based on models trained on different embeddings. Random forest classifiers were trained to predict glioblastoma subtypes (AC-like, OPC-like, NPC-like, MES1, MES2) using GNN embeddings, MOFA embeddings, or raw spatial transcriptomics (ST). GNN embeddings outperformed MOFA in both accuracy and macro F1-scores and achieved accuracy and F1-score close to ST, which represent an upper bound given that subtype labels were derived from ST-based deconvolution. b) Probability regression of subtype composition using neural networks. Raw ST features achieved the best performance as an upper bound (since labels were derived from ST deconvolution), while GNN embeddings substantially outperformed MOFA. c) Visualization of predicted subtype maps in a representative sample. GNN embeddings successfully recovered all five subtypes with coherent spatial organization, while MOFA produced strongly biased annotations, with very few MES2- and OPC-like spots. Raw ST features most closely matched the reference deconvolution-based annotations. d-f)

Unsupervised clustering and differential gene expression (DEG) analysis on the same representative sample. The left panel shows spatial visualization of unsupervised clusters derived from GNN embeddings, illustrating spatially coherent cluster boundaries consistent with tumor architecture. The right panels display volcano plots for three example clusters, highlighting significantly upregulated and downregulated genes; marker genes enriched in each cluster are highlighted.

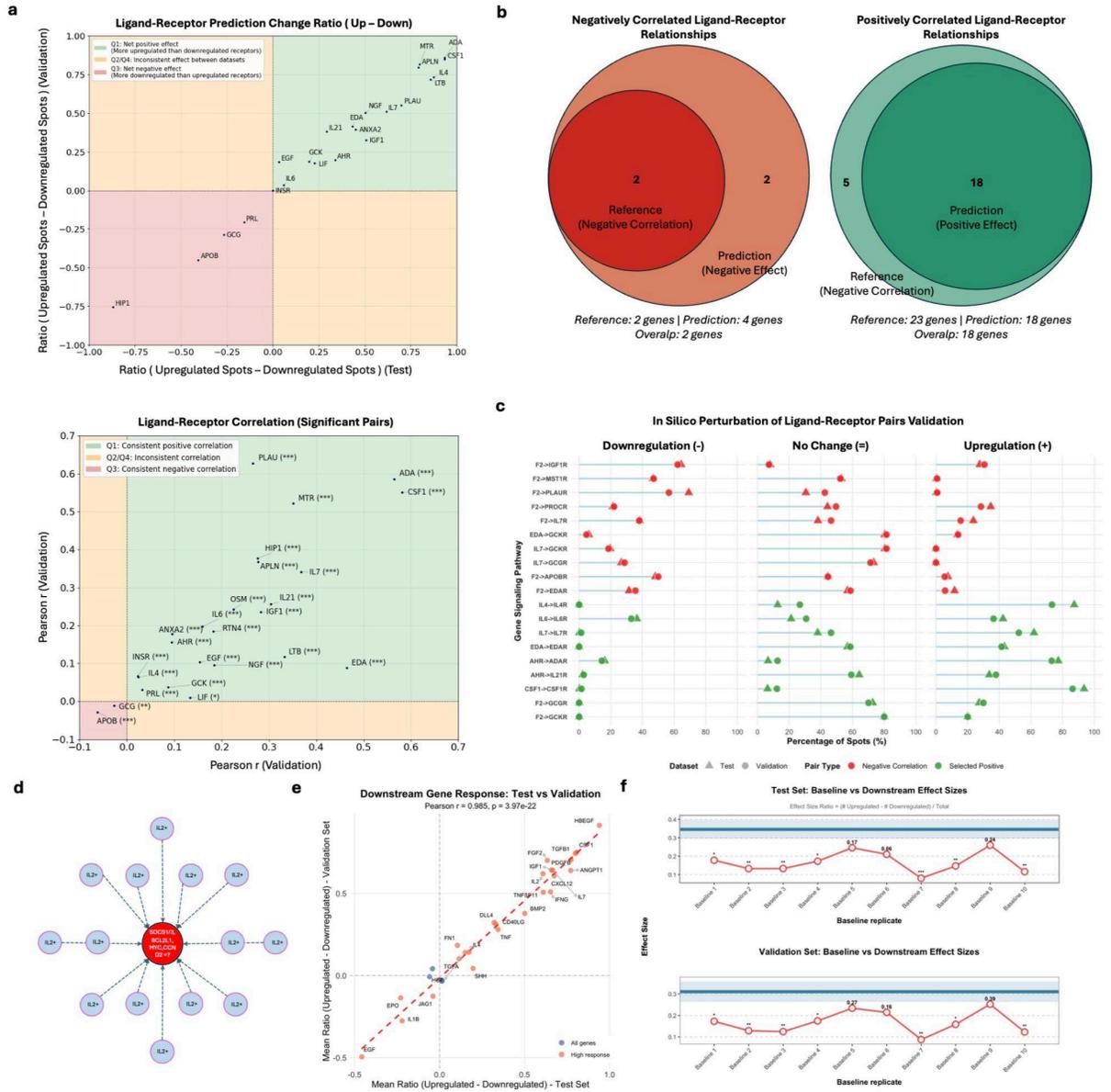

**Figure 5. Validation of spatial ligand–receptor regulatory relationships through *in silico* perturbation experiments.** a) Scatter plots showing ground-truth correlation and GNN predicted changes in expression of ligand–receptor gene pairs. Top: Predicted ligand–receptor gene expression changes under perturbation, comparing ratios of upregulated spots versus downregulated spots in the test set (X-axis) and validation set (Y-axis). The positive (green) and the negative (red) quadrants highlight ligand–receptor pairs with consistent effects on both validation and test sets. Bottom: Pearson correlation coefficients between the mean ligand gene expression in neighboring spots and receptor gene expression in central spots over subgraphs in test set (X-axis) and validation set (Y-axis). Ligand–receptor pairs with significant and consistent correlation directionality in the test and validation data are shown in positive (green) and negative (red) quadrants. b) Venn diagrams showing overlap between correlation-based reference and *in silico* perturbation GNN predictions. Left: Negative ligand–receptor relationships, where the GNN model correctly identified two ligand-receptor pairs. Right: Positive ligand–receptor relationships, where the GNN model correctly predicted 18 out of 23 reference pairs. c) Lollipop plots showing *in silico* perturbation validation for selected gene pairs. For each gene pair, plots display the proportion of central spots with predicted upregulation, downregulation, or no change in the receptor gene after simulated upregulation of the ligand gene in neighboring spots. Triangles represent test set predictions and circles represent validation set predictions. Green markers indicate positively correlated pairs (where ligand upregulation is expected to increase receptor expression based on correlation analysis), while red markers indicate negatively correlated pairs, where correlation was computed on the validation and test sets as in a). d) Schematic representation of downstream gene perturbation experiment design: central spots (red node) containing downstream target genes (SOCS1/3, BCL2L1, MYC, CCND2) and 14 neighboring spots. Input/ligand genes (IL2+, EGF+) are artificially perturbed in all neighboring spots, and the GNN model predicts the masked downstream gene expression in the central spot. e) Scatter plot showing predicted downstream gene expression changes under perturbation, comparing ratios of upregulated versus downregulated downstream genes in the test set (X-axis) and validation set (Y-axis). The ratio is computed as the proportion of upregulated minus downregulated downstream genes when comparing minimum (downregulation) versus maximum (upregulation) input gene perturbation levels. f) Line plots comparing effect sizes between downstream target genes and randomly selected baseline control genes across 10 random baseline replicates. Top: Test set. Bottom: Validation set. The blue horizontal line with shaded region shows the mean effect size for downstream genes (known regulatory targets) ± standard error. The red line with circular dots shows the effect size for baseline genes (randomly selected controls) across 10 baseline replicates under different random seeds. Effect size ratio is computed as (number of upregulated minus number of downregulated genes) divided by total, comparing maximum (upregulation) versus minimum (downregulation) input gene perturbation levels.